\documentclass[letterpaper, 10 pt, conference]{ieeeconf}  

\IEEEoverridecommandlockouts                              

\usepackage{graphics} 
\usepackage{epsfig} 
\usepackage{multirow}

\title{\LARGE \bf
Time Series Motion Generation \\Considering Long Short-Term Motion
}

\author{Kazuki Fujimoto$^{1}$, Sho Sakaino$^{2}$, and Toshiaki Tsuji$^{3}$
\thanks{$^{1}$Kazuki Fujimoto is a student with the Graduate School of Science and Engineering,
Saitama University, 255 Shimo-Okubo, Sakura-ku, Saitama City, Saitama 338-8570, Japan
       {\tt\small email: k.fujimoto.423@ms.saitama-u.ac.jp}}%
\thanks{$^{2}$Sho Sakaino is with the Graduate School of Systems and Information Engineering, University of Tsukuba, 1-1-1 Tennodai, Tsukuba, Ibaraki 305-8577, Japan and the JST PRESTO
       {\tt\small email: sakaino@iit.tsukuba.ac.jp}}%
\thanks{$^{3}$Toshiaki Tsuji is with the Graduate School of Science and Engineering,
Saitama University, 255 Shimo-Okubo, Sakura-ku, Saitama City, Saitama 338-8570, Japan
       {\tt\small email: tsuji@ees.saitama-u.ac.jp}}%
}
\begin{document}

\maketitle
\thispagestyle{empty}
\pagestyle{empty}
\begin{abstract}
Various adaptive abilities are required for robots interacting with humans in daily life. It is difficult to design adaptive algorithms manually; however, by using end-to-end machine learning, labor can be saved during the design process. In our previous research, a task requiring force adjustment was achieved through imitation learning that considered position and force information using a four-channel bilateral control.
Unfortunately, tasks that include long-term (slow) motion are still challenging. Furthermore, during system identification, there is a method known as the multi-decimation (MD) identification method. It separates lower and higher frequencies, and then identifies the parameters characterized at each frequency. Therefore, we proposed utilizing machine learning to take advantage of the MD method to infer short-term and long-term (high and low frequency, respectively) motion. In this paper, long-term motion tasks such as writing a letter using a pen fixed on a robot are discussed. We found differences in suitable sampling periods between position and force information. The validity of the proposed method was then experimentally verified, showing the importance of long-term inference with adequate sampling periods.
\end{abstract}
\section{INTRODUCTION}
Soon, robotic automation is expected to expand from use in factories as seen in recent years and into open environments. However, because conventional robots are programmed only to reproduce a designed trajectory, they cannot deal with changes in objects and environments. Hence, manual labor remains. For robots to adapt to such changes, it is necessary to program for each situation. This involves significant cost, time, and effort because the patterns to achieve tasks are infinite. For example, grasping an object can be challenging as there is the necessity to grasp with object stiffness and shape in mind. Some studies have attempted to solve this problem with hardware such as jamming grippers~\cite{jamming} and suction hands~\cite{suction hand}; however, the hardware can only grip matched objects (i.e., hardware physical characteristics restrict the scope of objects) furthering the problem. Therefore, studies to adapt to open environments using machine learning, which is software designed by data from many situations, have been attracting attention. Machine learning is now widely accepted because of its high generalization capability, which increases applications to robotics such as trajectory generation. However, designing machine learning is still complicated because features of robots and their surrounding environments must be selected by experts familiar with the tasks. Hence, so-called end-to-end learning has been a subject of research. This type of learning is known as end-to-end because you directly learn what you want from the inputs and generated outputs. Levine {\it et al.} succeeded in grasping multiple objects using reinforcement learning based on end-to-end learning~\cite{google}. Learning in this case required two months and 800,000 repetitions, resulting in poor practicality even when considering task difficulty. Therefore, recent studies have reported significantly reduced training data by imitating and learning human manipulation skills in a process referred to as imitation learning or learning from demonstration~\cite{LfD}~\cite{ogata}. Some studies demonstrated the ability to learn trajectory generation from position information~\cite{mit}~\cite{uc_berceley}, but there were not sufficient abilities to achieve various tasks, including force regulation, because each motion can be described as a combination of position and force controllers~\cite{analysis}. Additionally, Pacchierotti {\it et al.} reported that in a peg-in-a-hole experiment using a remote-control system, feeding reaction force back to the operator improved work efficiency~\cite{peg-in-hole}. Therefore, using force information is necessary for motion generation. Despite this, machine learning has rarely exploited both force information and force control using this fact. This is because action and reaction forces applied at the same place cannot be separated, hence making a force controller is difficult. In contrast, Yokokura {\it et al.} demonstrated that the force information needed to reproduce motion can be obtained using bilateral control~\cite{motion_copy1}, which is a remote-control technology using a master robot and a slave robot~\cite{bilateral}~\cite{micro-macro}. The master robot senses an action force from an operator, and the slave robot senses the reaction force from the environments. However, this method~\cite{motion_copy1} is intended to reproduce only the same behavior. Thus, the task cannot be achieved if the environment changes even slightly. In other words, such a method lacks robustness. Conventional methods of learning from force information are not clear; although, several methods on the effectiveness of using force information have been suggested~\cite{iit1}~\cite{iit2}. Recently we reported on the use of imitation learning for object manipulation that can implement position and force control using bilateral control~\cite{adachi}. Thanks to force control, our method has great generalization ability. The network model learned from the motion of drawing lines using a ruler, and the actual test was drawing a curve using a protractor. Note that it required only 15 trials to obtain training data. Although conventional methods are too slow, it is worth noting that the motion obtained by our method was as fast as human operation. This is because our method used training data to compensate for the phase delay of robots. However, the motion in our previous demonstration included only one action, and motions with multiple actions (long-term inference) still remain challenging. \par
\begin{figure}[t]
  \centering 
  \includegraphics[width=9.5cm]{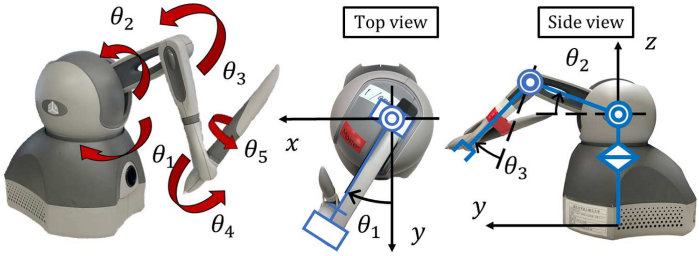}
  \caption{Definition of robot joints and coordinate system}
  \label{angle}
\end{figure}
Controlling robots involves a process of system identification in which dynamical systems' parameters are obtained from control inputs and measured outputs. In other words, parameters for determining robot motion are identified by end-to-end data. Therefore, there are similarities between machine learning and system identification, and we surmised that system identification methods would also be effective in machine learning. To the best of our knowledge, few studies have applied system identification methods to machine learning for robot motion generation; however, this is starting to attract attention in other research fields~\cite{learning_method for id1}~\cite{learning_method for id2}. Robotic physical dynamic characteristics spread from the low frequency region to the high frequency region and so on. Because the cost function in the usual least-squares method emphasizes high frequency regions results in poor identification accuracy in low frequency regions, Adachi {\it et al.} proposed the MD identification method~\cite{MD1}. This method makes it possible to identify low frequency region characteristics as accurately as those in the high frequency region. The method then identifies each frequency separately and connects the identified outputs. Another study also reported the effectiveness of the MD method~\cite{MD2}. Okamoto {\it et al.} reported that the MD method is effective at synthesizing speech waveforms~\cite{wavenet}. Because the low and high frequency regions are equivalent to long-term (slow) and short-term (fast) motions, respectively, the MD method will improve long-term inference in machine learning. Therefore, the conventional method~\cite{adachi} can be improved by integrating the MD method to consider long-term and short-term motion. In the proposed model, neural network inputs and outputs are designed using the MD method. Experimental results show that this proposed method's performance is improved but does not show satisfactory performance. Because of this, one can see that the position and force have different adequate sampling periods. Thus, we propose a method that considers long-term position information and short-term force information. Additionally, when considering the long-term information method, Yamashita {\it et al.} proposed the multiple timescales recurrent neural network (RNN) with hierarchical neuron counts with different response speeds~\cite{MTRNN} and has been refined and disseminated in many subsequent studies~\cite{MTRNNogata}. Instead, our proposal differs in that it only changes the sampling time at the input stage, making network implementation simple and reducing training costs compared with the multiple timescales RNN. The validity of the proposed method is experimentally verified, and a robot obtained the ability to write the letter A under different heights that were unknown in advance.\par
The remainder of this paper consists of the following sections. Section II explains the control system. Section III explains system identification. Section IV describes the training method. Section V demonstrates the proposed method's validity through experiments. Section VI concludes this paper and discusses future works.
\begin{figure}[t]
  \begin{center}
    \includegraphics[width=7.0cm]{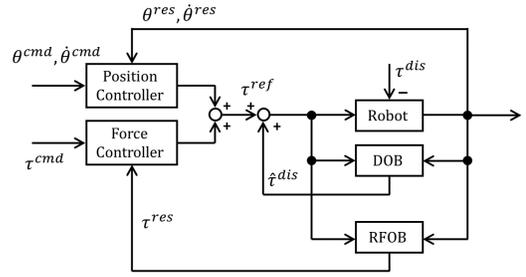}
    \caption{The manipulator's block diagram}
    \label{block_diagram}
  \end{center}
\end{figure}
\section{CONTROL SYSTEM}
\subsection{Manipulator}
We used two Geomagic Touch haptic devices manufactured by 3D Systems as manipulators (Fig.~\ref{angle}). Joint angles for the manipulators were as shown on the right side of Fig.~\ref{angle}. The fourth and fifth joints were fixed and did not move. Geomagic Touch devices can measure each joint's angle, and we calculated angular velocity using a pseudo derivative. $\hat{\tau_{dis}}$ represents the estimated disturbance torque value, which was calculated by a disturbance observer (DOB)~\cite{dob}. Additionally, reaction torque was calculated using a reaction force observer (RFOB)~\cite{rfob}. Detail of the RFOB are described in section~\ref{sub:id}. The manipulator system is shown as a block diagram in Fig.~\ref{block_diagram}. 
Here $\theta$, $\dot{\theta}$, and $\tau$ refer to joint angles, robot velocity, and robot torque, respectively. The superscripts ``res,'' ``ref,'' and ``cmd'' indicate response, reference, and command values, respectively, and the superscripts ``m'' and ``s'' indicate master and slave, respectively. 
The controller is composed of a combination of position and force controllers, with the position controller consisting of a proportional and derivative controller and the force controller consisting of a proportional controller.
\subsection{Four-channel bilateral control}
The angle and torque control targets are defined as follows:
\begin{eqnarray}
  \theta^{res}_{m}-\theta^{res}_{s} & = & 0 \label{eq1} \\
  \tau^{res}_{m}+\tau^{res}_{s} & = & 0 \label{eq2}.
\end{eqnarray}
In four-channel bilateral control, synchronization between master and slave angles is bidirectional, and at the same time, torque responses follow the law of action and reaction\cite{micro-macro}. Position and force controller properties are defined as follows:
\begin{eqnarray}
  \tau^{ref}_{s}&=&\frac{J}{2}(K_{p}+K_{v}s)(\theta^{res}_{m}-\theta^{res}_{s})-\frac{1}{2}K_{f}(\tau^{res}_{m}+\tau^{res}_{s}) \label{eq3} \nonumber \\
\end{eqnarray}
\begin{eqnarray}
  \tau^{ref}_{m}&=&\frac{J}{2}(K_{p}+K_{v}s)(\theta^{res}_{s}-\theta^{res}_{m})-\frac{1}{2}K_{f}(\tau^{res}_{s}+\tau^{res}_{m}). \label{eq4} \nonumber \\
\end{eqnarray}
Here, $K_{p}$ and $K_{f}$ represent the position and force control gain, respectively, and their values are shown in Table~\ref{gain}. $J$ indicates the identified inertia, and the control period was 1~msec.\par
\begin{table}[t]
  \begin{center}
    \caption{Gains of robot controller }
    \label{gain}
    \scalebox{0.8}[0.9]{ 
    \begin{tabular}{ccc} \hline
      \multicolumn{2}{c}{Parameter} & value \\ \hline \hline
      $K_{p}$ & Position feedback gain & 121.0 \\
      $K_{d}$ & Velocity feedback gain & 22.0 \\
      $K_{f}$ & Force feedback gain & 1.0 \\
      $g$ & Cut-off frequency of pseudo derivative [{\rm rad/sec}] & 40.0 \\ 
      $g_{DOB}$ & Cut-off frequency of DOB [{\rm rad/sec}] & 40.0 \\ 
      $g_{RFOB}$ & Cut-off frequency of RFOB [{\rm rad/sec}] & 40.0 \\ \hline \hline
    \end{tabular}
    }
  \end{center}
\end{table}
\subsection{Control system tuning} \label{sub:id}
Physical parameters of the robots were identified using a conventional identification method~\cite{yamazaki}. The actual reaction force can be estimated by subtracting friction and gravitational force from the estimated disturbance by the DOB~\cite{dob}~\cite{rfob}. In this paper, we calculated the reaction torques of each joint as follows:
\begin{eqnarray}
  \tau^{res}_{1} &=& \tau^{dis}_{1}-D\dot{\theta_{1}} \label{eq5} \\
  \tau^{res}_{2} &=& \tau^{dis}_{2}-g_{c1}\cos\theta_{2}-g_{c2}\sin\theta_{3} \label{eq6} \\
  \tau^{res}_{3} &=& \tau^{dis}_{3}-g_{c3}\sin\theta_{3} \label{eq7}.
\end{eqnarray}
~~Here the parameters $D$ and $g_{c}$ represent the friction compensation and gravity compensation coefficients, respectively. The identified parameters are shown in Table~\ref{id_parameter}.
\begin{table}[t]
  \begin{center}
    \caption{The identified system parameters}
    \label{id_parameter}
    \scalebox{0.8}[1.0]{ 
    \begin{tabular}{cccc} \hline
      \multicolumn{2}{c}{Parameter} & Master & Slave \\ \hline \hline
      $J_{1}$ & Joint~1's inertia [{\rm mNm}] & 4.20 & 4.45 \\
      $J_{2}$ & Joint~2's inertia [{\rm mNm}] & 5.58 & 5.26 \\
      $J_{3}$ & Joint~3's inertia [{\rm mNm}] & 1.51 & 1.63 \\
      $D$ & Friction compensation coefficient [${\rm mkg m^2/s}$] & 12.1 & 12.7 \\
      $g_{c1}$ & Gravity compensation coefficient~1 [{\rm mNm}] & 135 & 139 \\
      $g_{c2}$ & Gravity compensation coefficient~2 [{\rm mNm}] & 98 & 96 \\
      $g_{c3}$ & Gravity compensation coefficient~3 [{\rm mNm}] & 123 & 136 \\ \hline \hline
    \end{tabular}
    }
  \end{center}
\end{table}
\section{SYSTEM IDENTIFICATION}
System identification consists of following process flow:
\begin{enumerate}
  \item Design the identification experiment
  \item Perform the Identification experiment
  \item Pre-process the data \label{pre}
  \item Apply the system identification method.
\end{enumerate}
~~Here, decimation is included in the data pre-process step. Pre-processing consists of a low-pass filter (LPF), which can remove sensing noise, and down-sampling, which avoids overfitting in high frequency regions. The least-squares method is one of the most widely used system identification methods; however, the cost function weighs high frequency regions more than low frequency regions~\cite{MD1}\cite{MD2}. Although it is possible to prevent overfitting in the high frequency range by decimation, it varies depending on the frequency range to be identified. Thus, it is impossible to identify the entire region with high accuracy with a single instance of decimation. Therefore, an MD identification method using multiple decimation instances and LPFs was proposed.
\section{TRAINING METHOD}
\subsection{Neural networks model}
The network in the proposed method is comprised of an RNN, which is a neural network with a recursive structure where outputs of one neuron are utilized as inputs for other neurons. In other words, an RNN is a network that holds time series information. This network contributes to natural language processing and voice processing~\cite{RNN1}~\cite{RNN2}. Recently, an RNN was applied to robot operation~\cite{RNN3}. Long short-term memory (LSTM) is a special type of RNN that can learn long-term dependencies. It was first introduced in 1997~\cite{LSTM} and has been refined and disseminated in many subsequent studies.\par
The neural network used in this study is shown in Table~\ref{nn_structure}. The inputs were the slave robot's angle, angular velocity, and torque for each joint, meaning there were nine inputs as the robot has three joints. Similarly, the outputs were the angle, angular velocity, and torque for each joint of the master robot. To account for the RNN calculation time, it should be noted that the outputs inferred the states 20~msec after the inputs.
\begin{table}[t]
  \caption{Neural Network's structure}
  \label{nn_structure}
  \begin{center}
    \begin{tabular}{cccc}\hline
      Layer & Input & Output & Activation Function \\ \hline \hline
      1st & \multirow{2}{*}{9} & \multirow{2}{*}{50} & \multirow{2}{*}{tanh} \\
      (LSTM) & & & \\
      2nd & \multirow{2}{*}{50} & \multirow{2}{*}{50} & \multirow{2}{*}{tanh} \\
      (LSTM) & & & \\
      3rd & \multirow{2}{*}{50} & \multirow{2}{*}{9} & \multirow{2}{*}{identity mapping} \\
      (Linear) & & & \\ \hline \hline
    \end{tabular}
  \end{center}
\end{table}
\subsection{Normalization}
\begin{figure}[t]
  \begin{center}
    \includegraphics[width=8.5cm]{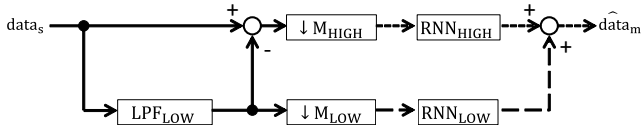}
    \caption{A block diagram of neural networks based on the MD method. First, input data was filtered by a designed LPF. Next, each frequency region's data are resampled to decrease to the desired sampling time. Third, the data is input to each ${\rm RNN}$. Finally, the outputs are added together.}
    \label{excute_model1}
  \end{center}
\end{figure}
\begin{figure}[t ]
  \begin{center}
    \begin{tabular}{c}
      \begin{minipage}{0.33\hsize}
        \begin{center}
          \includegraphics[clip, width=2.9cm]{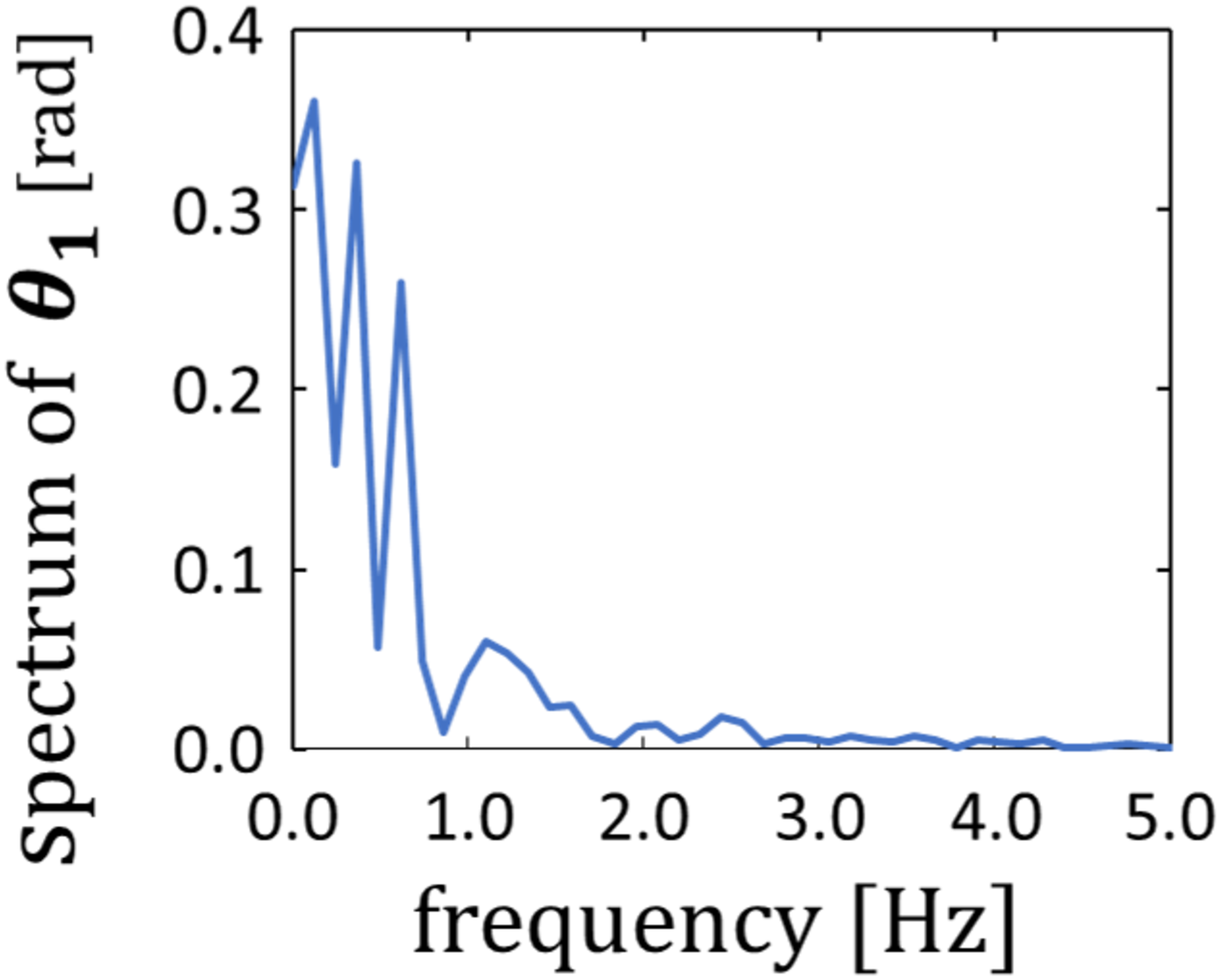}
          \hspace{1.5cm} (a) Spectrum of $\theta_{1}$
        \end{center}
      \end{minipage}
      \begin{minipage}{0.33\hsize}
        \begin{center}
          \includegraphics[clip, width=2.9cm]{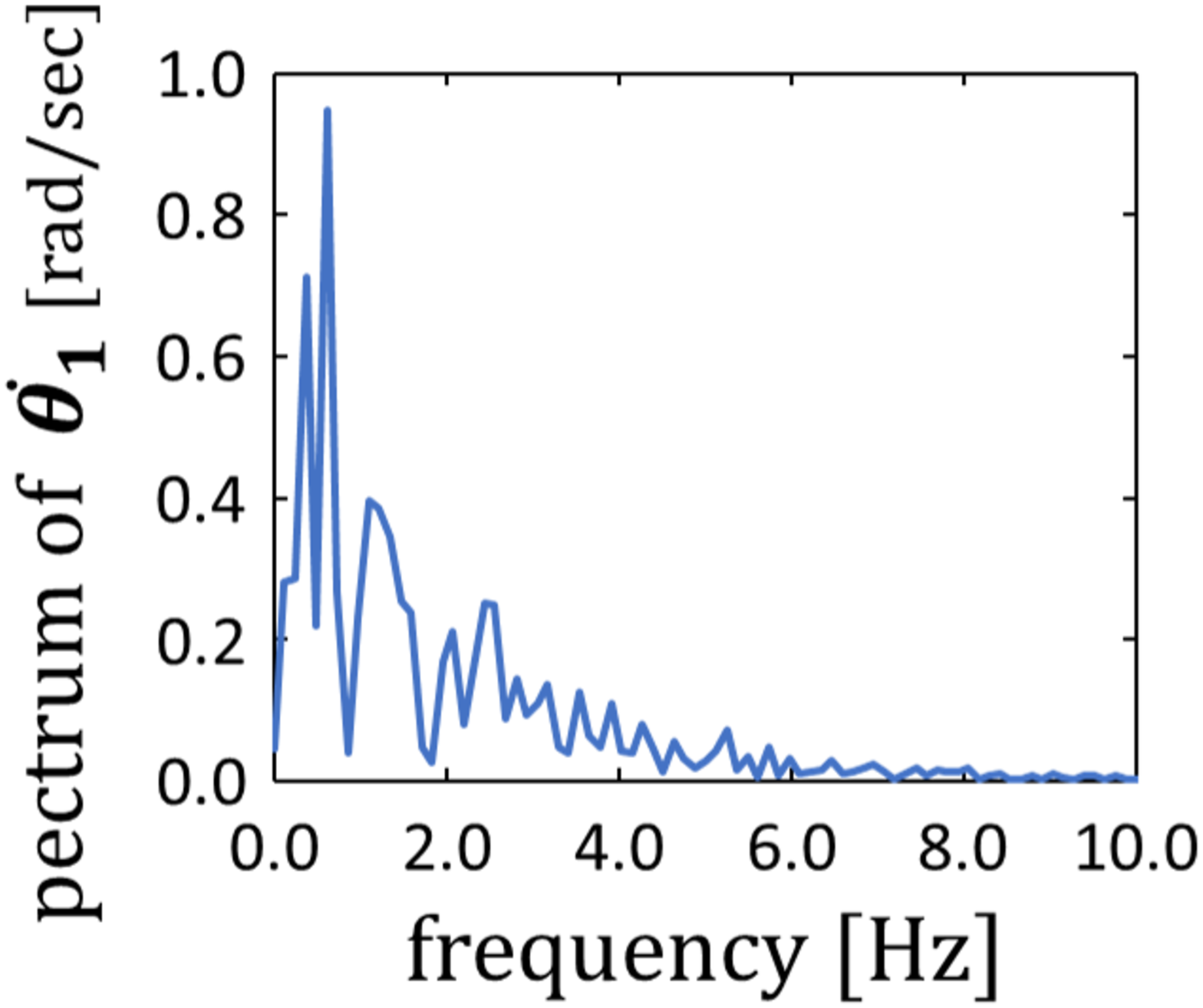}
          \hspace{1.5cm} (b) Spectrum of $\dot{\theta_{1}}$
        \end{center}
      \end{minipage}
      \begin{minipage}{0.33\hsize}
        \begin{center}
          \includegraphics[clip, width=2.9cm]{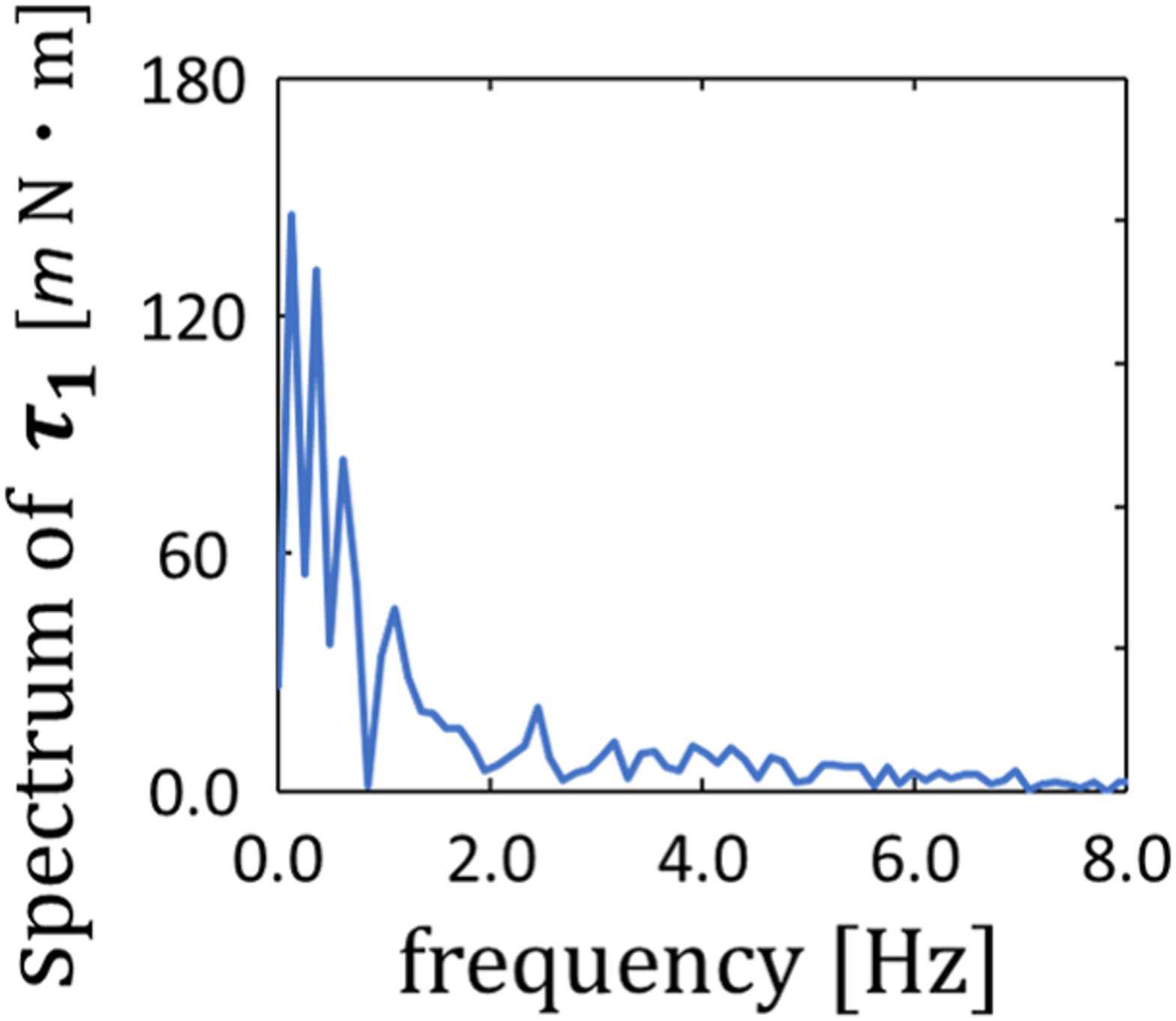}
          \hspace{1.5cm} (c) Spectrum of $\tau_{1}$
        \end{center}
      \end{minipage}
    \end{tabular}
    \caption{Spectrum of first joint.}
    \label{spectrum}
  \end{center}
\end{figure}
If the ranges of input and output data are significantly different because of unit system differences or other reasons, then scaling to equalize the data is effective.
This technique is known as normalization. Because the angle, angular velocity, and torque in this paper have different ranges, they must be normalized. Input data was normalized using a min-max normalization method, which is one of the most widely used normalization methods.
The normalization function is shown as follows:
\begin{eqnarray}
  d_n &=& \frac{d - d_{min}}{d_{max} - d_{min}} \label{eq8}.
\end{eqnarray}
~~Here, $d$ indicates raw data, and $d_{n}$, $d_{max}$, and $d_{min}$ represent normalized data, data maximum value, and data minimum value, respectively. Each of the maximum and minimum values is designed using the training data's maximum and minimum values. Then, output data was anti-normalized as follows:
\begin{eqnarray}
  d &=& d_n (d_{max} - d_{min}) + d_{min} \label{eq9}.
\end{eqnarray}
\subsection{Mini-batch}
In this paper we used a dataset consisting of 45 training data entries, each consisting of 750 samples.
The mini-batch process utilized for learning is shown as follows:
\begin{enumerate}
  \item Randomly select one data collection from the dataset \label{item1}
  \item Randomly select consecutive 300 data samples from the collection selected in \ref{item1}) \label{item2}
  \item Repeat \ref{item1}) and \ref{item2}) 100 times and regard it as a mini-batch. \label{item3}
\end{enumerate}
Unless otherwise noted the epoch number was 2000 and the model's learning time was approximately 35 minutes. In this paper, the computer used for this process had an Intel Core i7 CPU 32 GB memory, and a NVIDIA GTX 1080 Ti GPU.
\subsection{Proposed method}
In this subsection, we propose two methods: one based on the MD method and another that considers the result.
\subsubsection{Method based on MD method}
In reference to the MD method, the network was designed as shown in Fig.~\ref{excute_model1} (hereinafter, referred to as the MD model). $\downarrow{\rm M_{HIGH}}$ represents resampling to 20~msec, and $\downarrow {\rm M_{LOW}}$ indicates resampling to 400~msec. The ${\rm LPF_{LOW}}$ cut-off frequency was designed under the following sampling theorem:
\begin{eqnarray}
  g_{LOW} = \pi / st_{d} \label{eq10}.
\end{eqnarray}
~~Here, $g_{LOW}$ represents the cut-off frequency of ${\rm LPF_{LOW}}$, and $st_{d}$ represents the resampled sampling time. The sampling period in experiment was set to 400~msec to separate fast and slow motions, thus, ${g_{LOW}}$ of 400~msec was 7.85~rad/sec. ${\rm RNN_{HIGH}}$ and ${\rm RNN_{LOW}}$ then ran every 20~msec and 400~msec, respectively. The inputs and outputs of ${\rm RNN_{HIGH}}$ were high frequency datasets from the slave robot and master robot, respectively. Similarly, the inputs and outputs of ${\rm RNN_{LOW}}$ were low frequency datasets from the master and slave robots, respectively.\par
The angle, velocity, and force spectrum from the training data for the first joint (the slave robot) are shown in Fig.~\ref{spectrum}, which clearly shows that the angular spectrum existed in the low frequency region. Certainly, humans determine the trajectory of writing letters before they begin to write them, either planning the trajectory offline or very slowly online. Long-term inference should then be taken into consideration regarding position information. Therefore, this method was conducted to further clarify this consideration. Moreover, there is discussion in the RNN community that RNN performance cannot be sustained if there are differences in input and output between training data and the execution phase~\cite{model_error1}~\cite{model_error2}. In other words, input and output data are independently given as training data during training, whereas, during execution, output data have strong dependencies on previously input time series data making the situation different. In consideration of this, it is assumed that offline trajectory generation is better than online trajectory generation. In contrast, offline trajectory generation cannot handle sudden changes as much as motion copying can~\cite{motion_copy1}. Therefore, it is necessary to have a learning method with both the long-term estimation ability of offline trajectory generation and the immediate response capability of online generation. 
\subsubsection{Method considering suitable sampling periods for information}
\begin{figure}[t]
  \begin{center}
    \includegraphics[width=8.5cm]{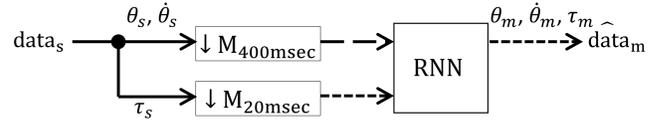}
    \caption{Block diagram showing a neural network that learns position and force information at different sampling rates.}
    \label{excute_model2}
  \end{center}
\end{figure}
\begin{figure*}[t]
  \begin{center}
    \includegraphics[height=7.0cm]{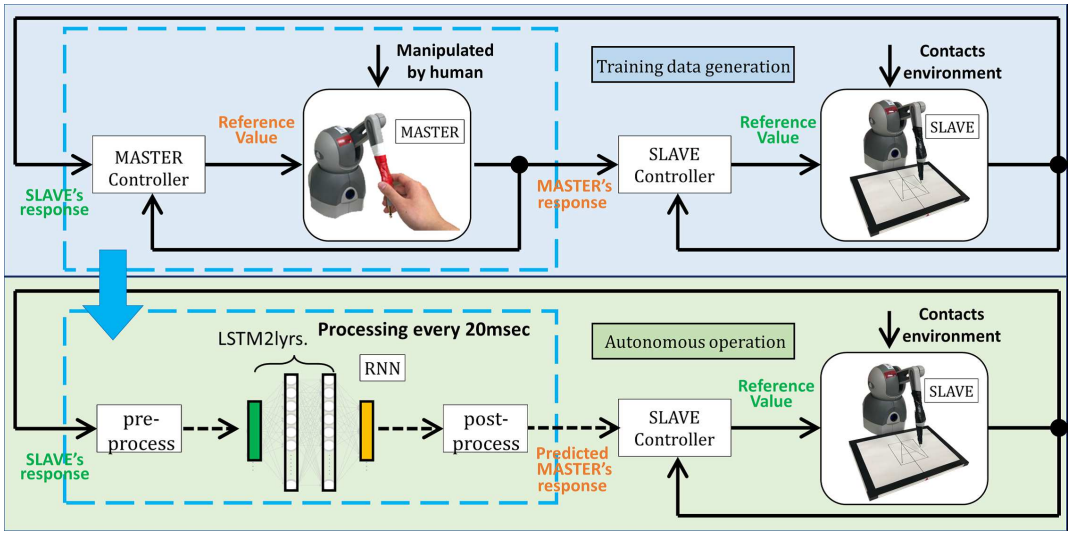}
  \end{center}
  \caption{Schematic of two stages; training data generation and execution. The top figure is the block diagram describing where robots are manipulated by an operator using four-channel bilateral control. Training data is generated based on this diagram. In contrast, the bottom figure represents the block diagram for when a slave robot executes tasks autonomously according to the trial model. In this stage, the top dashed square is replaced with the bottom dashed square. In other words, a neural network is necessary to learn human manipulation.}
  \label{stage}
\end{figure*}
\begin{figure}[t]
  \begin{center}
    \begin{tabular}{c}
      \begin{minipage}{0.33\hsize}
        \begin{center}
          \includegraphics[clip, width=1.8cm]{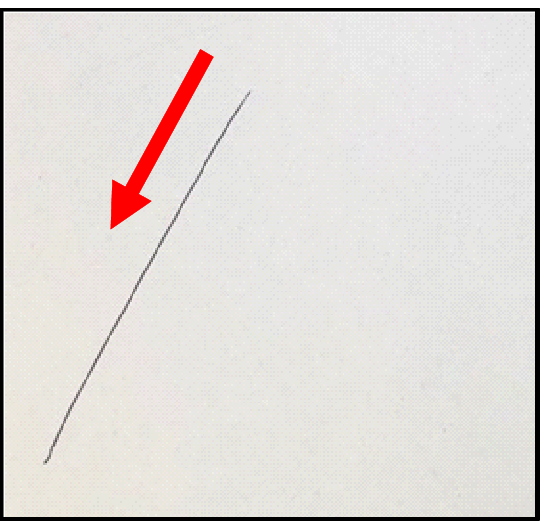}
          \hspace{1.6cm} (a) Step 1
        \end{center}
      \end{minipage}
      \begin{minipage}{0.33\hsize}
        \begin{center}
          \includegraphics[clip, width=1.8cm]{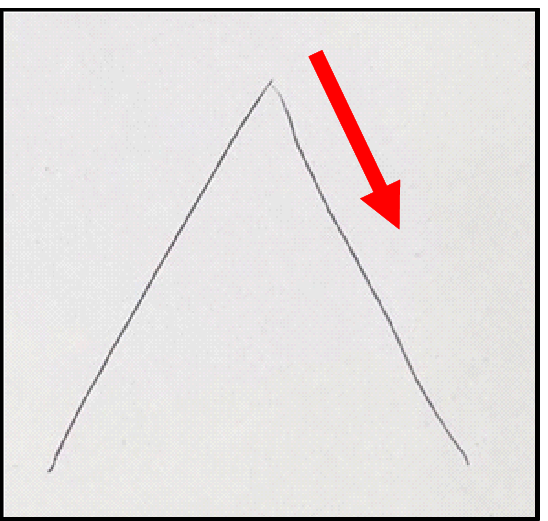}
          \hspace{1.6cm} (b) Step 2
        \end{center}
      \end{minipage}
      \begin{minipage}{0.33\hsize}
        \begin{center}
          \includegraphics[clip, width=1.8cm]{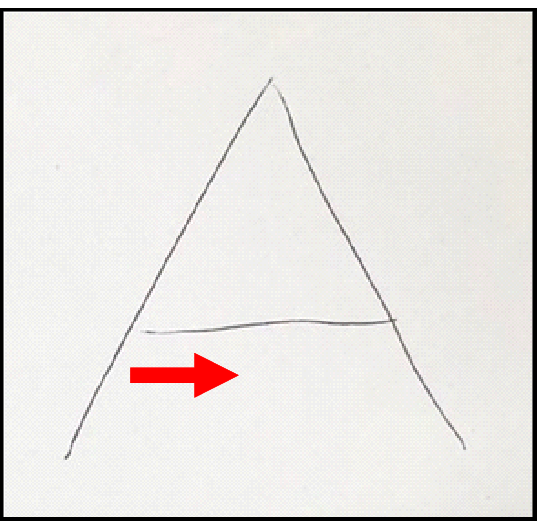}
          \hspace{1.6cm} (c) Step 3
        \end{center}
      \end{minipage}
    \end{tabular}
    \caption{Stroke order of training data for drawing the letter A. First, a line was drawn from the upper middle to the lower left. Next, a line was drawn from the upper middle to the lower right. Finally, a line was drawn from the left to the right in the middle.}
    \label{stroke_order}
  \end{center}
\end{figure}
Here we propose another method to learn both fast and slow motion, shown in Fig.~\ref{excute_model2} (hereinafter, referred to as PLT model, short for Position Long-Term). The network was designed for the ${\rm RNNs}$ to run every 20~msec, and the inputs and outputs were datasets from the slave and master robot, respectively. It is worth noting that regarding input, position information was updated every 400~msec, whereas force information was updated every 20~msec.
\section{EXPERIMENT}
\subsection{Generating training data}
\begin{figure}[t]
  \begin{center}
    \includegraphics[width=3.5cm]{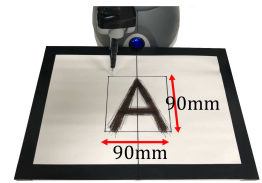}
    \caption{Completed writing on paper using training data. The letter A is written on a 90~mm square paper. This character was not written by the robot itself but rather an operator manipulated the robots to write the letter using bilateral control.}
    \label{alphabet}
  \end{center}
\end{figure}

The top of Fig.~\ref{stage} shows a block diagram of the training data generation stage. In this paper, the robot wrote a letter using a pen fixed to it as a task that required long-term motion. Training data was generated using four-channel bilateral control and saved every 1~msec. An operator manipulated the master robot, and the slave robot wrote the letter A. The operation is finally performed by the slave robot via a neural network that has learned to infer the master robot's response from the slave robot's response. The training data consisted of three steps shown in Fig.~\ref{stroke_order}. 
To learn how to write a letter even if the height of the paper changes, a robot wrote on paper with heights of 10~mm, 40~mm, and 70~mm. Because 15~sec of trial data was available for the training dataset every 20~msec, 750 pairs of input-output training data entries were generated. Then, trials of 15 secs at each height were performed 15 times for a total of 45 trials. The paper on which all training data were written is shown in Fig.~\ref{alphabet}. Generating training data required less than 30 minutes.
\begin{figure}[t]
  \begin{center}
    \includegraphics[width=9.0cm]{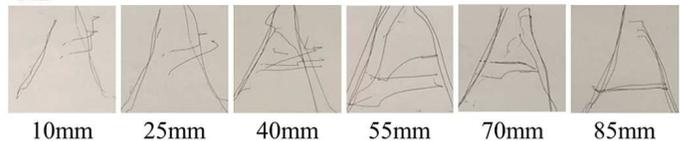}
    \caption{Results of experiment~1. The top of figures are the CONV. model and the others are the MD model. In both models, the robot ran until it wrote letters three times in the same place.}
    \label{result1}
  \end{center}
\end{figure}
\subsection{Experiment 1}
This experiment was carried out to verify the network's effectiveness in learning long-term (slow) and short-term (fast) motions separately based on the MD method. As a comparison, an ${\rm RNN}$ was designed to input slave robot data at every 20~msec and output master robot data 20 msec after inputs. This is referred to as the CONV. model~\cite{adachi}.\par
Experimental results of autonomous operations in the CONV. model and MD model are shown in Fig.~\ref{result1}. Both models were trained with three stage heights: 10~mm, 40~mm, and 70~mm. The situations with heights of 25~mm and 55~mm are unlearned interpolation, whereas the 85~mm scenario is extrapolation. As can be seen in Fig.~\ref{result1}, the CONV. model cannot write the letter A at all heights. In contrast, although it is not perfect, the letter written using the MD model captures the characteristics of the letter and demonstrates that the use of long-term information using the MD method is effective.\par
\begin{figure}[t]
  \begin{center}
    \includegraphics[width=6.5cm]{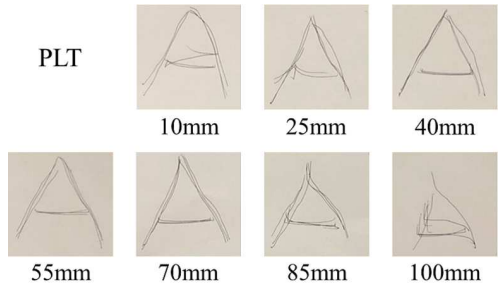}
    \caption{Results of experiment~2. The robot ran until it wrote letters three times in the same place.}
    \label{result2}
  \end{center}
\end{figure}
\begin{figure}[t]
  \begin{center}
    \includegraphics[width=8.5cm]{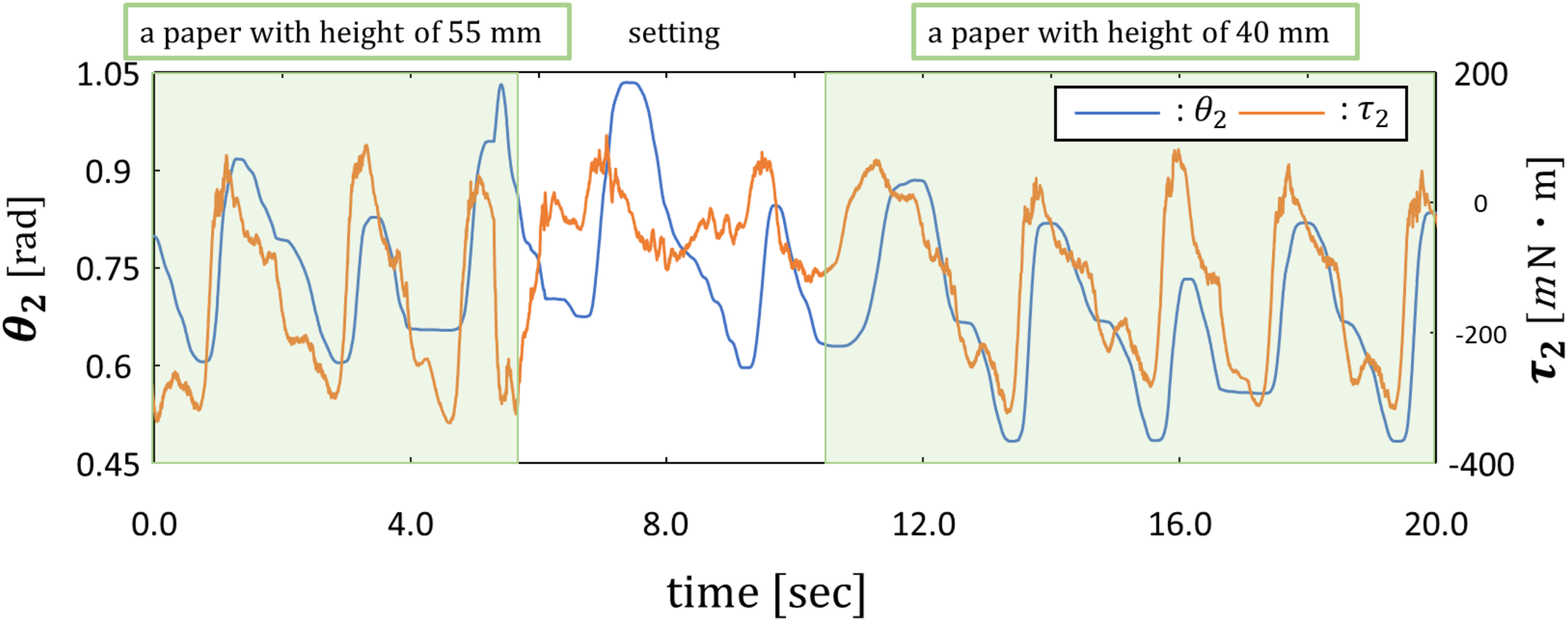}
    \caption{Transition of $\theta_{2}$ and $\tau_{2}$. The figure shows the angle of writing and the response value of torque. The left side shows a paper height of  55~mm and the right side shows a paper height of 40~mm.}
    \label{graph}
  \end{center}
\end{figure}
\subsection{Experiment 2}
Results of autonomous operations using the PLT model are shown in Fig.~\ref{result2}, which clearly shows that the letter A can be written even on unlearned heights. Despite neural networks being vulnerable to extrapolation, the letter could be written even at a height of 85~mm. However, at a height of 100~mm, it was impossible to write the letter clearly unlike at the other paper heights. Additionally, Fig.~\ref{graph} shows an experimental result in which the paper height was suddenly changed during the writing process. The output angle then changed, but the output torque scarcely changed. Considering these results, the PLT model learned to write letters with a similar writing pressure for each paper height. 
\section{CONCLUSION}
In this paper, we focused on high frequency and low frequency regions during motion generation and proposed two RNN models for learning that consider both long-term and short-term motion. To verify the two models, a task of writing a letter using a pen fixed on a robot was given. In the MD model inspired by the MD identification method, the robot could not write the equivalent characters from the training data. The task's angular spectrum existed in the low frequency region, and humans plan writing trajectory offline or very slowly online, thus, it is necessary that the model have suitable sampling periods for position and force information. Thus, we designed the PLT model, which combined long-term inference capabilities in trajectory generation with short-term inference abilities and could write letters on a piece of paper of unlearned heights. The PLT model has the feature of delaying position information input and updating it. Although the multiple timescales RNN is another method for considering long-term information, our proposal is different in that it changes the sampling time during the input stage. It is then assumed that our model's learning is easier than the multiple timescales RNN. Additionally, this model suppresses the vibration caused by sequential prediction of position information. Unfortunately, the way to determine an adequate sampling time is still an open problem.\par
Future work is needed to apply the proposed method to other tasks and automate the design using the proposed PLT model. Furthermore, the cause of the proposed method's robust behavior will be investigated.
\section*{ACKNOWLEDGMENT}
This work was supported by JST PRESTO Grant Number JPMJPR1755, Japan.

\end{document}